\pdfoutput=1
\documentclass[11pt]{article}

\usepackage{EMNLP2023}
\usepackage{times}
\usepackage{latexsym}

\usepackage[T1]{fontenc}
\usepackage[utf8]{inputenc}

\usepackage{microtype}

\usepackage{xcolor}
\usepackage{colortbl}
\usepackage{inconsolata}
\usepackage{comment}
\usepackage{xspace}
\usepackage{multirow}
\usepackage{multicol}
\usepackage{graphicx}
\usepackage{booktabs}
\usepackage{pgfplots}
\usepackage{pgfplotstable}
\usepackage{booktabs}
\usepackage{todonotes}
\usepackage{tikz}
\usetikzlibrary{patterns}
\usepackage{arydshln}
\usepackage{tablefootnote}
\usepackage{soul}
\usepackage{subfigure}
\usepackage{amsmath}
\usepackage{listings}
\usepackage{graphicx}
\usepackage{array}
\usepackage{pifont} % for \ding{51} and \ding{55}
\usepackage{scalerel,xparse}
\definecolor{darkgreen}{RGB}{0,200,0}
\NewDocumentCommand\emojipie{}{
    \scalerel*{
        \includegraphics[scale=0.5]{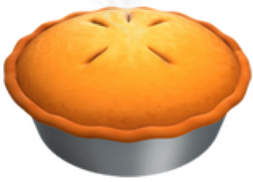}
    }{X}
}
\newcommand{\authorspace}{\hspace{0.3cm}}

\definecolor{darkgreen}{RGB}{0,200,0}
\definecolor{lightblue}{RGB}{211,227,253}
\newcommand{\model}{\textsc{TART}\xspace}
\newcommand{\dataset}{\textsc{ToolTab}\xspace}
\title{TART\emojipie{}{}: An Open-Source Tool-Augmented Framework for Explainable Table-based Reasoning}

\author{
    \bf Xinyuan Lu$^1$ \authorspace
     \bf Liangming Pan$^2\thanks{~~Corresponding Author}\:\, $  \authorspace
    \bf Yubo Ma$^3$ \authorspace
    \bf Preslav Nakov$^4$ \authorspace 
    \bf Min-Yen Kan$^1$ \authorspace
  \\
  \vspace{0.1mm}
 $^1$National University of Singapore 
 $^2$University of Arizona\authorspace \\
 $^3$Nanyang Technology University
  $^4$MBZUAI \\
  {\tt luxinyuan@u.nus.edu} \authorspace 
  {\tt liangmingpan@arizona.edu} \authorspace
  {\tt yubo001@e.ntu.edu.sg} \authorspace \\
  {\tt preslav.nakov@mbzuai.ac.ae} \authorspace
  {\tt kanmy@comp.nus.edu.sg} 
}

\begin{document}
\maketitle
\begin{abstract}
Current Large Language Models (LLMs) exhibit limited ability to understand table structures and to apply precise numerical reasoning, which is crucial for tasks such as \textit{table question answering} and \textit{table-based fact verification}. To address these challenges, we introduce our \textit{Tool-Augmented Reasoning framework for Tables} (\model), which integrates LLMs with specialized tools. \model contains three key components: a \textit{table formatter} to ensure accurate data representation, a \textit{tool maker} to develop specific computational tools, and an \textit{explanation generator} to maintain explainability. We also present the \dataset dataset, a new benchmark designed specifically for training LLMs in table--tool integration. Our experiments indicate that \model achieves substantial improvements over existing methods (\textit{e.g.,} Chain-of-Thought) by improving both the precision of data processing and the clarity of the reasoning process. Notably, \model paired with \texttt{CodeLlama}  achieves 90.0\% of the accuracy of the closed-sourced LLM \texttt{GPT-3.5-turbo}, highlighting its robustness in diverse real-world scenarios. 
Both code and data are openly available at \url{https://github.com/XinyuanLu00/TART}. 
\end{abstract}

\section{Introduction}

Tabular data is prevalent across multiple fields such as scientific research, financial reporting, and healthcare records~\cite{DBLP:conf/ijcai/0001CHZZZLHZ22}. Manual handling of such data can be both routine and error-prone, or may require specialized skills, highlighting the need for automated table reasoning to improve efficiency~\cite{DBLP:journals/tacl/Badaro0P23}. 
% improve efficiency and accessibility~\cite{DBLP:journals/tacl/Badaro0P23}. 
Typical table-based reasoning tasks include \textit{ table question answering} (TQA), which extracts precise information from tables to answer given queries~\cite{DBLP:hybridqa,DBLP:tatqa,DBLP:journals/corr/abs-2104-00369} and \textit{table-based fact verification} (TFV), which verifies the correctness of statements by cross-referencing them with facts stored in tables~\cite{DBLP:conf/semeval/WangMDR21,lu2023scitab}. 

\begin{figure}[!t]
    \centering
    \includegraphics[width=7.7cm]{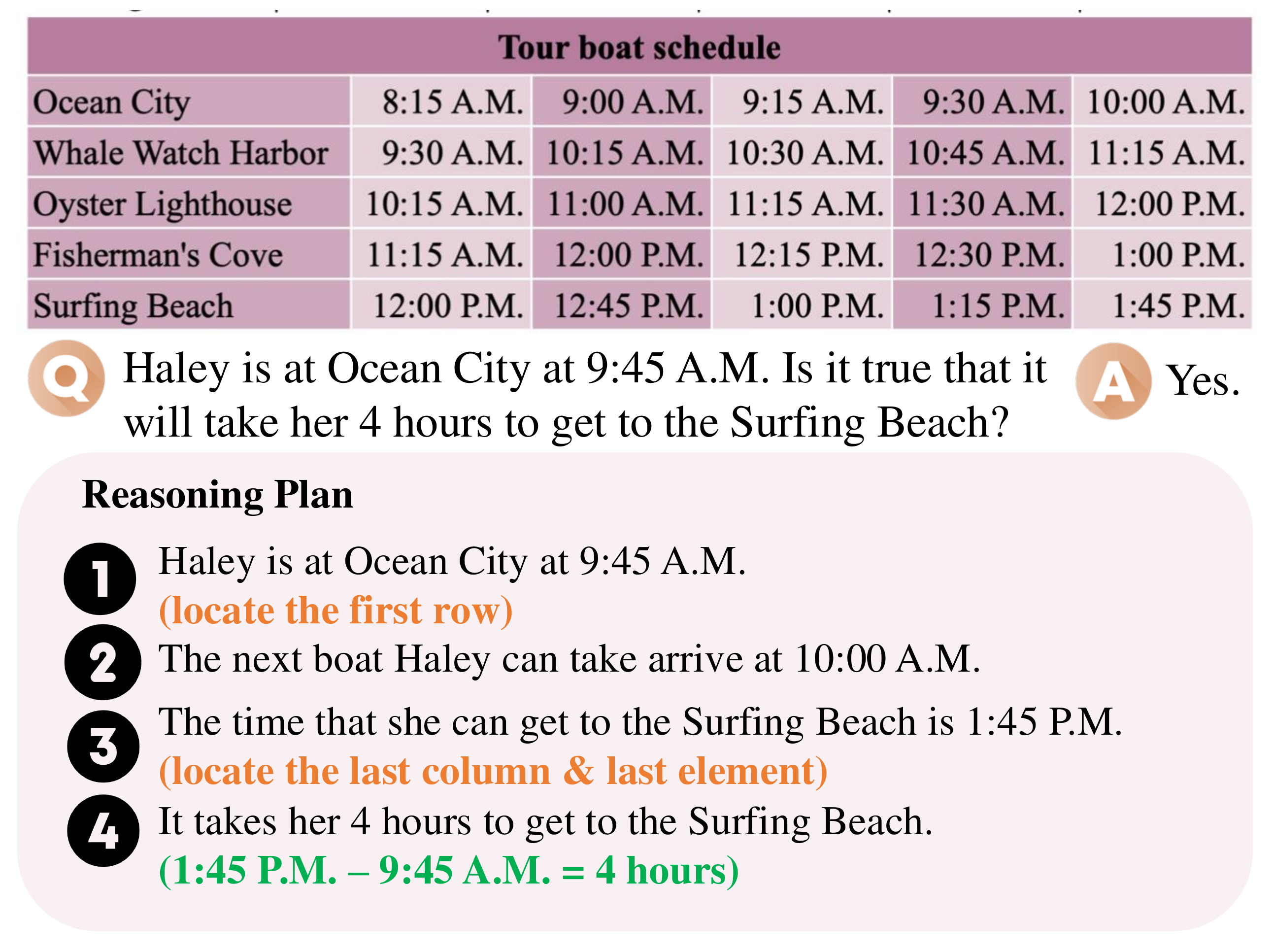}
    \caption{Example of the TableQA task, demonstrating the verification of travel time via boat schedule and the critical skills needed for accurate table reasoning: table structure understanding, precise numerical calculations, and executing sequential reasoning steps.}
    \label{fig:example}  
   \vspace{-0.5cm}
\end{figure}

\begin{figure*}[!ht]
    \centering
    \includegraphics[width=16cm]{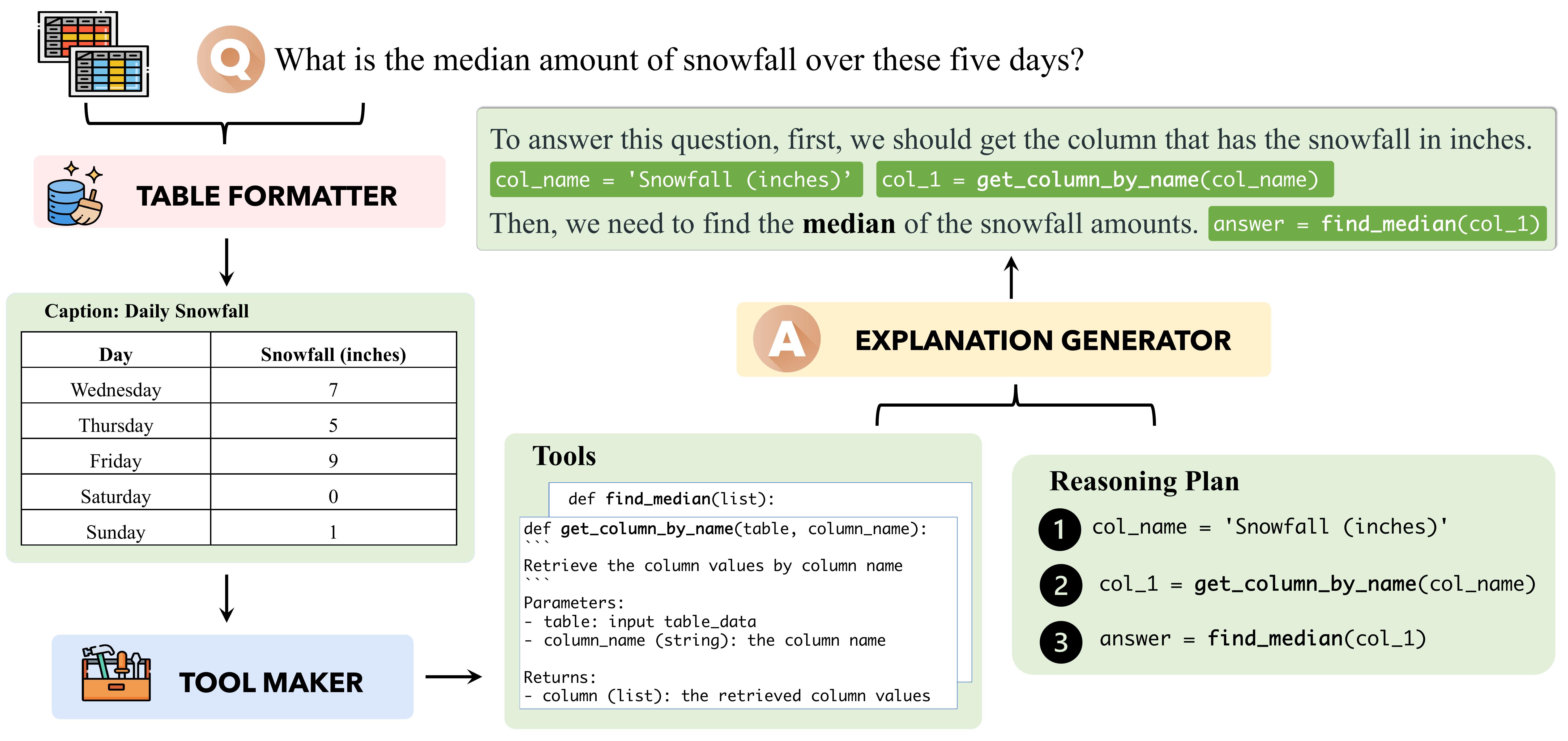}
    \caption{An overall framework of \model, which contains three main modules: (\emph{i})~\textit{table formatter}, which prepares and organizes the raw table data, (\emph{ii})~\textit{tool maker}, which creates specialized tools for precise table manipulation, and (\emph{iii})~\textit{explanation generator}, which produces user-friendly explanations integrating the output of different tools.}
    \label{fig:model}
    \vspace{-0.5cm}
\end{figure*}

%Related work and Gap:
Modern large language models (LLMs) such as GPT-4~\cite{DBLP:journals/corr/abs-2303-08774} have shown remarkable reasoning capabilities across a variety of tasks, spurring interest in their application to table-based tasks~\cite{ye2023large}. However, table-based reasoning presents unique challenges for LLMs, which are primarily trained on sequential text data~\cite{tablellama}, as illustrated by a real-world example in Figure~\ref{fig:example}. \textit{(1) Understanding and operating on table structure}: LLMs must adapt to the non-linear format of the tables, which demands unique reasoning skills such as recognizing headers, interpreting the roles of the rows and the columns, and precisely extracting information from relevant table cells. \textit{(2) Precise numerical reasoning:} Tables often contain quantitative information that requires precise calculations and comparisons. LLMs must perform operations such as summation, averaging, or trend analysis accurately, often over multiple cells or tables, which is a shift from their usual text-based reasoning tasks~\cite{DBLP:conf/acl/HerzigNMPE20,DBLP:conf/iclr/LiuCGZLCL22}. \textit{(3) Reasoning planning:} LLMs often need to plan several reasoning steps ahead.  This includes decomposing the original question, determining relevant table parts, and anticipating intermediate calculations or data transformations. All three challenges manifest in Figure~\ref{fig:example}.

Existing approaches that use LLMs for table-based reasoning can be broadly classified into two categories. One is \textit{chain-of-thought (CoT) reasoning}~\cite{DBLP:CoT}, in which the model is prompted to perform step-by-step reasoning over the input table flattened as a textual sequence~\cite{tablellama,tabcot,chen2023large,ye2023large}. Despite its flexibility, CoT often lacks precision in tabular operations and numerical reasoning, such as sorting, counting, and filtering~\cite{protrix}. The second approach, \textit{program-based reasoning (PoT)}~\cite{gao2023pal,chen2022program} prompts the model to generate SQL or Python code to enable precise reasoning~\cite{rethinkingtabular,reactable,chainoftable,protrix}. However, this method struggles with reasoning planning and its reasoning is less understandable to humans~\cite{reactable}. Therefore, there is potential value in integrating the advantages of program-based and textual reasoning, to achieve both high precision and explainability in table-based reasoning. 

%% Purpose
Inspired by the recent paradigm of tool-augmented language models~\cite{DBLP:toolsurvey,schick2023toolformer}, we propose \textit{\textbf{T}ool-\textbf{A}ugmented \textbf{R}easoning framework for \textbf{T}ables} (\model), which integrates external tool calling into the chain-of-thought reasoning process, as shown in Figure~\ref{fig:model}. 
% \model operates through a series of interconnected modules: 1)
Initially, \model processes the input table using a specialized module \textit{table formatter} to clean and to format the raw table data, preparing it for the subsequent table operations. Subsequently, the \textit{tool maker} calls specialized tools (Python functions) for tabular manipulation and numerical reasoning (e.g., adding columns, selecting rows, and grouping). Alongside these tools, \model also crafts a \textit{reasoning plan} that outlines the programmatic calling sequence of the tools, specifying the necessary arguments and the expected return values for each call. 
% The reasoning plan acts as a blueprint, detailing which tools to use, the order of operations, and how to interpret the results. 
Finally, following the structured reasoning plan, the \textit{explanation generator} produces a hybrid text-and-program output, integrating calls to external tools into coherent, human-readable chain-of-thought explanations. In doing so, \model efficiently delegates table operations and precise numerical calculations to generated tools while preserving CoT's planning ability and explainability. 

To train the modules in \model, we further synthesize the \dataset dataset by distilling knowledge from a teacher LLM. 
We evaluate \model on nine different table-based reasoning benchmarks. The results highlight the effectiveness of integrating task-specific tools for enhancing complex reasoning capabilities. Notably, TART consistently outperformed the CoT baseline, achieving near-parity with \texttt{GPT-3.5-turbo} on benchmarks, showcasing its usefulness in real-world scenarios. 

In summary, our contributions are threefold:

$\bullet$ We propose \model, a novel framework that enhances table-based reasoning by integrating tools into the reasoning process, which addresses the limitations of current LLMs in handling table structures and executing precise calculations. 

$\bullet$ We develop \dataset, a comprehensive benchmark specifically designed to train LLMs on table-tool integration. It includes diverse real-world tables, uniform format, and careful validation to ensure high-quality training. 

$\bullet$ Our experiments confirm that \model not only improves the precision and the explainability of table-based reasoning, but also generalizes effectively to out-of-domain datasets.

\section{Related Work}
%new
Table-based reasoning tasks involve interpreting and manipulating data from structured tabular sources to answer questions, verify facts, or generate summaries. Early approaches used executable SQL or SPARQL to interact with tabular data~\cite{yin2016neural,yu-etal-2018-spider}, or graph neural networks to better encode table structures~\cite{zhong2020logicalfactchecker,yang-etal-2020-program}. However, they typically suffer from poor generalization capabilities due to their reliance on specific table formats and linguistic patterns. 
 
Recent advances in large language models (LLMs) have demonstrated significant potential in this area. Pre-training strategies that align LLMs with sentence-table pairs~\cite{DBLP:conf/iclr/ChenWCZWLZW20,DBLP:conf/acl/HerzigNMPE20,zhou2022tablebased,DBLP:conf/emnlp/GuF0NZ022,ye2023large,grounding} have improved table reasoning capabilities, while frameworks like TAP4LLM~\cite{tab4llm} optimizes table representations through sampling, augmentation, and serialization. Other works, such as \textit{ReAcTable}~\cite{reactable} and Chain-of-Table~\cite{chainoftable}, have introduced hybrid or explicit reasoning mechanisms to better integrate tabular data into reasoning chains. These methods primarily rely on textual reasoning strategies, such as chain-of-thought, which often lack the precision necessary for table manipulations and numerical reasoning. Efforts have also been made to evaluate and enhance LLM capabilities for table-related tasks. Studies like Table Meets LLM~\cite{tabmeetsllm} and Text2Analysis~\cite{text2analysis} introduce benchmarks for tasks such as cell lookup, row retrieval, and Python-based advanced data analysis, emphasizing challenges in table serialization and query understanding. Meanwhile, TabularNet~\cite{tabularnet} proposes novel architectures, combining graph-based and relational representations to improve semantic understanding of tabular data. However, these approaches typically focus on either enhancing representation or evaluation rather than augmenting reasoning precision. In addition to general-purpose frameworks, domain-specific solutions such as EHRAgent~\cite{ehragent} have been developed for multi-tabular reasoning in specialized domains like electronic health records (EHRs). While EHRAgent integrates domain-specific metadata and debugging strategies, its focus contrasts with \model, which is designed as a domain-agnostic solution for table reasoning across diverse contexts. Similarly, API-Assisted Code Generation~\cite{apiassistcodegeneration} translates queries into Python programs leveraging fixed APIs, differing from \model’s dynamic tool-augmented reasoning approach.

To address the limitations of prior work, \model extends the use of LLMs with integrated external tools, enabling precise table manipulations and numerical reasoning while maintaining explainability. By combining strategies such as table formatting for better representation, tool-based function execution for precision, and LLM-based reasoning for interpretability, \model advances the state of table-based reasoning tasks.

\section{Methodology}

Generally, a table-based reasoning model, $f_{\theta}(\cdot)$, parameterized by $\theta$, takes an input query $Q$ and a table $\mathcal{T}$ to produce a response $Y = f_{\theta}(Q, \mathcal{T})$. Based on this generic formulation, the nature of $Q$ and $Y$ differs depending on the specific table reasoning task: 

in table-based QA, $Q$ is a question and $Y$ is the answer; in table-based fact verification, $Q$ is a claim and $Y$ is its veracity label.
% ; in table summarization, $Q$ specifies the requirement and $Y$ is the summary
 A table $\mathcal{T}$ is characterized by a caption $P$ and its contents ${T_{i,j} \mid i \leq R_T, j \leq C_T}$, where $R_T$ and $C_T$ represent the numbers of rows and columns, respectively. Each cell $(i, j)$ contains data $T_{i,j}$. 

To build an accurate and explainable table-reasoning framework, our proposed \model
% \textit{\textbf{T}ool-\textbf{A}ugmented \textbf{R}easoning framework for \textbf{T}ables (\model)} 
integrates the call to external tools into the chain-of-thought reasoning process. \model consists of three reasoning modules (Figure \ref{fig:model}): 
\textcircled{\small{\textbf{1}}} \textit{Table Formatter}; \textcircled{\small{\textbf{2}}} \textit{Tool Maker}; \textcircled{\small{\textbf{3}}} \textit{Explanation Generator}.

\paragraph{1. Table Formatter.} \model first transforms the original table $\mathcal{T}$ with guidance from the query $Q$ into a formatted table $\mathcal{T}'$. The formatter optimizes data formats, aligns columns, and adjusts data types as needed for the query, producing a well-formatted table that is used in subsequent reasoning. 

\paragraph{2. Tool Maker.} Given % the cleaned table 
$\mathcal{T}'$ % and the query $Q$
, the \textit{tool maker} generates a set of candidate tools $S$ useful for solving $Q$. %the query. 
It also develops a reasoning plan $R$ that details the high-level reasoning which includes the tool calling order, as well as the necessary arguments and the expected return values for the tool calls. 
% $\mathcal{M}$ 

\paragraph{3. Explanation Generator.} Given the reasoning plan $R$ as a programmatic guide for chain-of-thought reasoning, the explanation generator is responsible for producing a user-friendly explanation $E$ that incorporates the use of the tools. The explanation also concludes with the final answer $\mathcal{A}$, derived from the reasoning plan $R$'s execution. 

\subsection{Table Formatter}
\label{sec:tableformatter}

We first train a specialized open-sourced large language model as the table formatter $\mathcal{F}$, which transforms the noisy raw input table $\mathcal{T}$, into a more structured and manageable format, $\mathcal{T}'$, to facilitate subsequent reasoning:
% \begin{equation*}
% \setlength{\abovedisplayskip}{1pt}
% \setlength{\belowdisplayskip}{1pt}
    $\mathcal{T}' = \mathcal{F}(\mathcal{T}, Q)$
% \end{equation*}
% \noindent 
where the output table $\mathcal{T}'$ is formatted according to three aspects. 1) \textit{Data Cleaning}: the model formats the cell values, such as removing currency symbols and textual footnotes to facilitate the execution of external functions to perform table operations or numerical reasoning. 2) \textit{Data Standardization} converts different data representations into a uniform format;\textit{e.g.,} transforming the data from ``\texttt{MM/DD/YYYY}'' to a consistent ``\texttt{YYYY-MM-DD}'' format across the entire table. 3) \textit{Error Handling}: the model also fixes obvious errors or missing values in the table, such as automatically inferring header names for columns without the table headers. 

We introduce the table formatter to ensure that the data in the input table is uniform and optimized for subsequent reasoning, especially to make it more compatible with function execution. In practice, we transform the formatted table $\mathcal{T}'$ into a Python array, facilitating easier interpretation and processing by subsequent reasoning modules. 

\subsection{Tool Maker}
\label{sec:toolmaker}

Recent studies have shown that LLMs have the capability of synthesizing relevant tools by understanding the problem context and creating solutions based on the crafted tools~\cite{schick2023toolformer,DBLP:conf/iclr/Cai00CZ24,DBLP:toolsurvey}. Motivated by this, we train another specialized LLM $\mathcal{M}$ as a \textit{tool maker}, which takes as input the reformatted table $\mathcal{T}'$ and the query $Q$ to generate a set of candidate tools $S$ and develops a reasoning plan $R$ that details the high-level reasoning steps: 
%\begin{equation*}
%\setlength{\abovedisplayskip}{1pt}
    $S, R = \mathcal{M}(\mathcal{T'}, Q).$
%\end{equation*}
The tool set $S = \{ s_1, \cdots, s_n \}$ consists of $n$ specialized tools, where each tool $s_i$ is a Python function that performs table operations (\textit{e.g.}, \texttt{get\_column\_by\_name}), numerical reasoning (\textit{e.g.}, \texttt{average}, \texttt{argmax}), or higher-level functions (\textit{e.g.}, \texttt{linear\_regression}). 
% In table-based reasoning, 
These automated tools are essential to handle reasoning tasks that textual-based LLMs cannot address effectively.  
% significantly enhancing their problem-solving capacities. 

Unlike previous work that manually defined a small number ($<10$) of hand-crafted tools~\cite{DBLP:conf/nips/LuPCGCWZG23,DBLP:conf/acl/PanWLLWKN23} or retrieved tools from a predefined set~\cite{DBLP:conf/iclr/QinLYZYLLCTQZHT24,DBLP:sciagent}, we choose to train a specialized \textit{tool maker} model that \textbf{learns to \textit{generate} tools dynamically, based on the specifics of the table and the context of the problem.} 
% the table and the problem context. 
% This maintains the ability of the model to ``extract'' tools that have seen in training from its parametric memory while also giving the flexibility for the model to \textit{invent} unseen tools when nessesory when encountering novel problems. 
This approach not only preserves the model's ability to ``extract'' previously encountered tools from its parametric memory, but also empowers the model to create novel tools as needed for unique problems, as shown in Section~\ref{sec:tool use}.
% Addressing long-tail problem
While generating tools offer greater flexibility, it is crucial to prevent the tool maker from creating overly-specific tools (\textit{e.g.}, \texttt{count\_people\_on\_third\_floor}), as this would hinder its ability to generalize to new problems. To address this issue, we incorporate \textit{tool abstraction} and \textit{tool deduplication} steps when constructing synthetic data for training the module (Section~\ref{subsec:model_training}). 

% Reasoning Plan
% In addition to generating useful tools, 
The model also constructs a high-level reasoning plan $R = [ r_1, \cdots, r_N ]$, which outlines how tools should be applied. The reasoning plan is formulated as a sequence of $N$ function calls. Each \textit{function call} $r_i = (s_i, A_i, V_i)$ includes the function $s_i \in \mathcal{S}$, the \textit{argument} $A_i$ passed to the function, and the \textit{variable} $V_i$ that stores the result of the function call $s_i(A_i)$. 

This reasoning plan acts as a programmatic blueprint, guiding the table-based reasoning process. Both the tool set $S$ and the reasoning plan $R$ are then provided to the explanation generator, producing the final explainable reasoning output. 

\subsection{Explanation Generator}
\label{sec:expgen}

While program-based reasoning plans are precise, they are often difficult for non-expert users to understand. Moreover, certain types of reasoning, such as commonsense or narrative-based reasoning, are better communicated in natural language. To address this, \model incorporates a specialized module called the \textit{explanation generator} $\mathcal{E}$, which generates chain-of-thought natural language explanations integrated with function calls, following the steps outlined in the reasoning plan $R$:
%\begin{equation*}
% \setlength{\abovedisplayskip}{1pt}
% \setlength{\belowdisplayskip}{1pt}
    $O = \mathcal{E}(S, R).$
%\end{equation*}
% \noindent 
The final output $O$ of \model provides detailed explanations for the function calls. For example, the function call \texttt{get\_column\_by\_name} is explained as, ``\textit{First, retrieve the column listing snowfall in inches.}'' Additionally, the explanation generator groups related function calls together to create coherent and easy-to-follow explanations, as illustrated in Figure~\ref{fig:model}.

\subsection{Model Training}
\label{subsec:model_training}

As no prior work adopts the tool-augmented LLM framework for table reasoning, there does not exist training data to train the modules \textit{Table Formatter}, \textit{Tool Maker} and \textit{Explanation Generator}
($\mathcal{F}$, $\mathcal{M}$, and $\mathcal{E}$) in \model. Previous studies have demonstrated that smaller LLMs can learn from distilling the generated outputs of larger teacher LLMs that have better reasoning capabilities~\cite{DBLP:conf/naacl/WestBHHJBLWC22,DBLP:conf/acl/WangKMLSKH23,DBLP:Husky}. Following this, we use a teacher LLM $\mathcal{L}$ to first synthesize tool-integrated solution trajectories for a set of seed table-based reasoning tasks. These high-quality solution trajectories serve as the blueprint from which we automatically extract and rearrange their components to build training sets for $\mathcal{F}$, $\mathcal{M}$, and $\mathcal{E}$. 

% How to choose datasets. what datasets we choose? 
% For tool maker, how to address over specific tools? 
% We name this dataset as TacTool
\paragraph{Training Data Synthesis.} For all modules, we use GPT-4 as the teacher LLM $\mathcal{L}$ to generate training data. As shown in Table~\ref{tab:dataset comparison}, we select five diverse table reasoning datasets: two from TQA and three from TFV, spanning general knowledge (Wikipedia) as well as domains such as finance, health, and scientific documents. These datasets provide a broad range of reasoning types. We few-shot prompt $\mathcal{L}$ to generate tool-integrated solutions for training instances for each dataset. In each solution, the model is prompted to clean the table, invent tools, and propose a reasoning plan with explanations. We provide our prompt in Appendix~\ref{append:prompt}.

After generating the solutions, we evaluate the final answers against the ground truth, retaining only the instances with correct answers. Subsequently, we refine the solutions by removing overly specific tools through \textit{tool abstraction} and \textit{tool deduplication}. Tool abstraction filters out tools that appear only once, keeping those with broader applicability. Tool deduplication consolidates similar tools that perform the same function, but have different names or implementations (e.g., \texttt{add} and \texttt{sum}). 
As a result, we obtain 11,701, 9,916, and 9,916 training instances for the table formatter $\mathcal{F}$, tool maker $\mathcal{M}$, and explanation generator $\mathcal{E}$, We refer to this training dataset as \dataset, with detailed statistics provided in Table~\ref{tab:training samples}.

\paragraph{Training Configurations.}
Instruction fine-tuning~\cite{DBLP:conf/acl/MishraKBH22,DBLP:journals/corr/abs-2210-11416} has emerged as a critical strategy that directs LLMs to adhere to specified instructions, facilitating their reasoning capability across a wide range of table-based tasks. Therefore, we use open-source LLMs with instruction tuning as the backbone models for the modules of \model, specifically \texttt{Llama-2-7b}~\cite{touvron2023llama}, \texttt{Llama-3-8b}, \texttt{CodeLlama-7b}~\cite{DBLP:journals/codellama} and \texttt{Deepseek-Coder-7b-Instruct-V1.5}~\cite{deepseek}. We fine-tune all \model modules independently on their respective training datasets from \dataset, using the standard next-token prediction objective.

\begin{table*}[!t]
    \centering
    \resizebox{0.95\textwidth}{!}{
\begin{tabular}{cllcccccc}
    \toprule
  & &&  \multicolumn{3}{c}{\textbf{TableFV}}&  \multicolumn{2}{c}{\textbf{TableQA}}\\   
    & \multicolumn{1}{c}{\textbf{Model}} & \textbf{Setting} & \textbf{TabFact} & \textbf{PubHT} & \textbf{SCITAB} & \textbf{TabMWP} & \textbf{FinQA} & \textbf{Avg. Acc.} \\
    \midrule
    \multirow{2}{*}{\uppercase\expandafter{\romannumeral1.}}
         & \multirow{2}{*}{\textbf{TableLlama}} & w/o Fine-tuning & 72.3 & 72.5 & 67.4 & 46.8 & 3.2 & 52.4 \\
         & & w/ DirectQA & 72.9 & 70.5 & \underline{74.2} & 48.4 & 3.7 & 54.0 \\
       % \cdashline{2-9}
       %   & \multirow{2}{*}{\textbf{StructLM}} & w/o Fine-tuning & 81.9 & 81.9 & \underline{78.1} & 73.9 & 10.6 & 65.3 \\
       %   & & w/ DirectQA & \underline{83.5} & 81.2 & 75.3 & 74.5 & 9.6 & 64.8 \\
    \midrule
    \multirow{9}{*}{\uppercase\expandafter{\romannumeral2.} }
         & \multirow{3}{*}{\textbf{Llama2-7b}}& w/ DirectQA &64.4&81.2&64.0&55.3&6.4&54.3 \\
         & & w/ CoT & 52.6 & 55.0 & 42.7 & 74.5 & 4.2 & 45.8 \\
     \rowcolor{lightblue}   & & w/ TART & 69.2 & 55.0 & 53.4 & 88.8 & 19.2 & 57.1 \textcolor{red}{\small{(+24.7\%)}}\\
       \cdashline{2-9}
                & \multirow{3}{*}{\textbf{Llama3-8b}}& w/ DirectQA&\underline{74.5}&\textbf{85.9}& \textbf{82.0}&68.6&10.6&64.3\\         
         & & w/ CoT & 48.4 & 62.4 & 41.0 & 88.3 & 8.5 & 49.7 \\
        \rowcolor{lightblue}  & &w/ TART & 69.7 & 68.5 & 47.2 & 92.6 & 27.1 & 61.0 \textcolor{red}{\small{(+22.7\%)}} \\
       \cdashline{2-9}
         & \multirow{3}{*}{\textbf{CodeLlama-7b}}& w/ DirectQA & 65.4&75.8&64.6&44.7&4.3&51.0 \\  
         & &w/ CoT & 45.2 & 51.7 & 38.8 & 70.7 & 2.7 & 41.8 \\
        \rowcolor{lightblue}  & & w/ TART & 66.5 & 69.8 & 44.9 & 90.1 & 25.0 & 59.3 \textcolor{red}{\small{(+41.9\%)}} \\
       \cdashline{2-9}
         & \multirow{3}{*}{\textbf{DeepSeek-7b}}&w/ DirectQA & 72.9&76.5&73.0&62.2&9.0&58.7 \\        
         & &w/ CoT & 52.1 & 62.4 & 45.5 & 84.6 & 8.5 & 50.6 \\
       \rowcolor{lightblue} & & w/ TART & 71.3 & 69.1 & 47.8 & \underline{93.1} & 30.9 & 62.4 \textcolor{red}{\small{(+23.3\%)}} \\
    \midrule
    \multirow{2}{*}{\uppercase\expandafter{\romannumeral3.} }
         & \textbf{GPT-3.5-turbo} & w/ TART & 78.7 & 63.6 & 59.3 & 88.3 & \underline{56.4} & \underline{69.3} \\
         & \textbf{GPT-4} & w/ TART & \textbf{87.7} & \underline{84.1} & 63.6 & \textbf{98.3} & \textbf{68.5} & \textbf{80.4} \\
    \bottomrule
    \end{tabular}
    }
    \caption{Performance evaluation across backbone models using the \model framework, highlighting the best (bold) and second-best (underlined) results. The accuracy is calculated on testing sets, with overall average accuracy in the last column (\textit{Avg. Acc.}). The red numbers indicate the average increase percentage over the CoT methods.}
    \label{tab:main results}
        \vspace{-0.3cm}
\end{table*}

%%%%EXPERIMENTS
\section{Experiments}
\paragraph{Datasets and Baselines.}
\label{sec:baselines}

To evaluate \model, we select two categories of benchmarks for table-based reasoning. \textit{(1) Table question answering (TQA)}: \texttt{WikiTableQuestion (WTQ)}~\cite{DBLP:wikitablequestion} focuses on simple factoid questions. \texttt{HiTab (HIT)}~\cite{DBLP:hitab}, \texttt{TabMWP (TMP)}~\cite{lu2023dynamic} and \texttt{FinQA (FQA)}~\cite{DBLP:journals/corr/abs-2109-00122} datasets focus on numerical reasoning reasoning. \texttt{TAT-QA (TAT)}~\cite{DBLP:tatqa} and \texttt{HybridQA (HYQ)}~\cite{DBLP:hybridqa} require joint reasoning over the table and the text for financial reports and Wikipedia tables, respectively. \textit{(2) Table-based fact verification (TFV)}: We select \texttt{TabFact (TAF)}~\cite{DBLP:conf/iclr/ChenWCZWLZW20}, \texttt{SCITAB (SCT)}~\cite{lu2023scitab}, and \texttt{PubHealthTab (PHT)}~\cite{akhtar-etal-2022-pubhealthtab} datasets, which focus on verifying facts based on tables from Wikipedia, scientific articles, and public health articles, respectively. % During testing, we use the publicly available test split. 

For baseline comparisons, we select well-known table-based open-source LLMs such as \texttt{TableLlama}~\cite{tablellama} 
% and \texttt{StructLM}~\cite{DBLP:structlm}
, as well as text-pretrained models (\texttt{Llama2-7b}, \texttt{Llama3-8b}) and code-pretrained models (\texttt{CodeLlama-7b}, \texttt{DeepSeek-Coder-7b}). We choose the 7b and the 8b versions to represent a balance between computational efficiency and the capacity for complex reasoning and generalization. For each model, we fine-tune with two settings: (1) \texttt{DirectQA}, where models generate answers directly from questions and tables, and (2) \texttt{Chain-of-Thought (CoT)} reasoning, which requires models to formulate a step-by-step reasoning process before concluding with an answer. 
% We use accuracy as the evaluation metric. 

\paragraph{Implementation.}
\label{sec:experiment implementations}
For TART, we use the answer given by executing the reasoning plan; if the reasoning plan is not executable, we use the answer given by CoT. 
For each model, we train the model with \dataset while leaving the rest (WTQ, HIT, TAT, and HYQ) as held-out unseen datasets. 
% For each model, we train the model on TMP, FQA, TAF, SCT, PHT datasets while leaving WTQ, HIT, TAT, and HYQ as held-out unseen datasets. 
All experiments were conducted on a GPU server with Intel Xeon Platinum 8480C (224) @ 2.900GHz CPU and 8 NVIDIA H100 (80G) GPUs. The training process for \texttt{Llama-2-7b-hf}, \texttt{CodeLlama-7b-hf}, and \texttt{deepseek-coder-7b-instruct-v1.5} requires a single GPU for approximately 20 hours, using a batch size of 4, learning rate of 5e-5, sequence length of 1500, gradient accumulation steps of 2, and 10 training epochs. Training \texttt{Llama-3-8b} required up to two GPUs for around 20 hours with the same settings. To minimize randomness, a temperature of 0.0 was used, while all other hyperparameters for sampling the output from the LLMs remained at their default values. For the closed-source version of \model, we use \texttt{GPT-3.5-turbo} and \texttt{GPT-4} with two in-context examples. 

\subsection{Main Results}
\label{sec: main results}

We first evaluate \model and the baselines on in-domain datasets, where their training sets are used to construct \dataset. The experimental results, as shown in Table~\ref{tab:main results}, reveal a notable performance improvement in our model compared to baseline models. We have four major observations. 

\vspace{0.1cm}

1. \model consistently outperforms \texttt{CoT} across all four backbone models and datasets. For example, with \texttt{CodeLlama-7b} as the backbone model, \model outperforms \texttt{DirectQA} and \texttt{CoT} by 16.3\% and 41.9\% on average, respectively. This highlights the effectiveness of integrating task-specific tools in enhancing complex reasoning capabilities. 

\vspace{0.1cm}

2. With \texttt{CodeLlama-7b} as the backbone model of \model, it achieves the highest accuracy increase of 41.9\%., whereas \texttt{Llama3-8b} shows the least improvement of 22.7\%. This discrepancy is likely because of \texttt{CodeLlama-7b}'s specialized pre-training in coding tasks, which enhances the capabilities of creating tools for structured queries and operations. 

\vspace{0.1cm}

3. The performance gains of \model also vary for different datasets, with \texttt{FinQA} showing the highest increase, while \texttt{PubHealthTab} shows the least. This discrepancy suggests that the financial focus of the \texttt{FinQA} dataset, which demands extensive numerical reasoning and structured data manipulation, benefits significantly from the \model approach. 

\vspace{0.1cm}

4. Using closed-source models (\texttt{GPT-3.5-turbo} and \texttt{GPT-4}) as the backbone models for \model achieves an average accuracy of 74.9, significantly outperforming the open-source counterparts, which average at 60.0 accuracy. Nonetheless, the highest-performing open-source model, \texttt{DeepSeek-7b}, reaches up to 90.0\% of GPT-3.5-turbo's performance and 77.6\% of \texttt{GPT-4}, illustrating the competitiveness of open-source models in the creation and use of tools despite the apparent model size gap. 

\begin{table}
    \centering
    \fontsize{8}{10}\selectfont
    \setlength{\tabcolsep}{1.3mm}
    \resizebox{0.48\textwidth}{!}{
\begin{tabular}{ccccccc}
\toprule
\textbf{Model}  & \textbf{TAF} & \textbf{PHT}  & \textbf{SCT}&\textbf{TMP}  & \textbf{FQA}& \textbf{Avg. Acc.}\\
\midrule
Llama2                       & 69.2 & 55.0 & 53.4 & 88.8 & 19.2 & 57.1                \\ 
\rowcolor{lightblue} - TabFT    & 67.7 & 39.1 & 50.4 & 67.8 & 17.2 & 48.4 \textcolor{red}{(-15.2\%)}  \\ \hdashline

Llama3                           & 69.7 & 68.5 & 47.2 & 92.6 & 27.1 & 61.0                \\
\rowcolor{lightblue} - TabFT     & 69.2 & 58.9 & 46.3 & 65.4 & 13.8 & 50.7 \textcolor{red}{(-16.9\%)} \\
\hdashline
CodeLlama                      & 66.5 & 69.8 & 44.9 & 90.1 & 25.0 & 59.3                \\ 
\rowcolor{lightblue} - TabFT   & 66.6 & 64.9 & 36.6 & 69.1 & 20.2 & 51.5 \textcolor{red}{(-13.2\%)}  \\ \hdashline
DeepSeek                       & 71.3 & 69.1 & 47.8 & 93.1 & 30.9 & 62.4                \\
\rowcolor{lightblue} - TabFT  & 58.0 & 58.8 & 45.0 & 71.3 & 16.7 & 50.0 \textcolor{red}{(-19.9\%)} \\ 
\bottomrule                                          
\end{tabular}
    }
    \caption{The ablation study of Table Formatter (\textit{- TabFT}) in \model with different backbone models. The red numbers indicate the average accuracy (\textit{Avg. Acc.}) drop percentage without Table Formatter for each model.}
   \vspace{-0.3cm}
    \label{tab:ablation study on table formatter}
\end{table}

\begin{figure*}[!th]
    \centering
    \includegraphics[width=2.1\columnwidth]{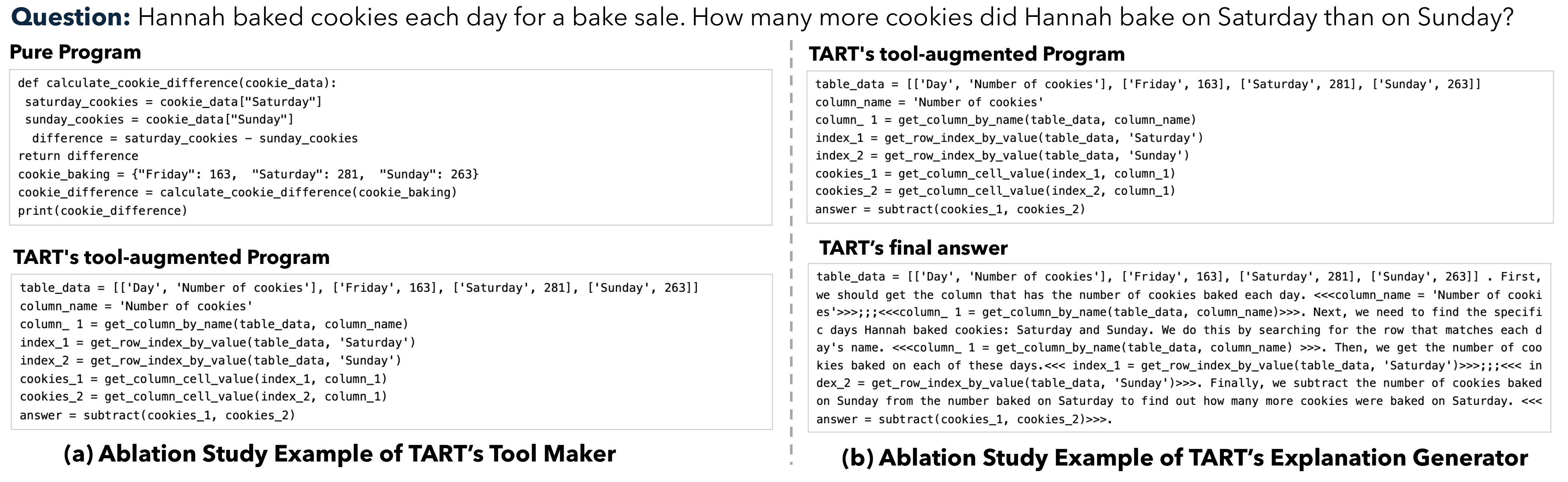}
    \caption{Ablation Study of \model 's \textit{Tool Maker} and \textit{Explanation Generator}. Panel (a) compares the pure program with \model's tool-augmented program, highlighting the effectiveness of Tool Maker. Panel (b) compares \model's tool-augmented program with \model's final answer, demonstrating the usefulness of Explanation Generator.}
    \label{fig:ablation_study_case}
    %\vspace{-0.3cm}
\end{figure*}

% Min: this is a narration of the results, but doesn't give any insight.  Can you remove the last sentence and tell us anything more specific?
\subsection{Ablation Study}
\paragraph{Ablation of the Table Formatter.}
% To evaluate the impact of the Table Formatter module, 
We conduct ablation experiments across all backbone models (Table~\ref{tab:ablation study on table formatter}). TART without the Table Formatter led to significant  and uniform performance drops of over 10\%. 
% Min: details are in the table anyways.  Don't waste the space.  
% 15.2\%, 16.9\%, 13.2\%, and 19.9\% for \texttt{Llama2-7b}, \texttt{Llama3-8b}, \texttt{CodeLlama-7b}, and \texttt{DeepSeek-7b}, respectively. 
This clearly demonstrates the effectiveness of the Table Formatter module in improving reasoning capabilities by ensuring consistent table representations.

\paragraph{Ablation of the Tool Maker.}
% If the Tool Maker module is removed, 
In this ablation, the framework directly generates programs instead of creating modular tools (Figure~\ref{fig:ablation_study_case}a). While this approach achieves functionality, it impacts both performance and explanability. These tools (e.g., \texttt{get\_column\_by\_name}) can be used repeatedly comparing to the tools that are overly specific (e.g.,\texttt{calculate\_cookie\_difference}).

\paragraph{Ablation of the Explanation Generator.}
The Explanation Generator module does not directly impact the final performance in terms of accuracy, but enhances the explanability of the outputs. For example, in Figure~\ref{fig:ablation_study_case}(b), \model's final answer provides more human-readble explanations, comparing to the program.

\begin{table}
    \centering
    \fontsize{8}{10}\selectfont
    \setlength{\tabcolsep}{1.3mm}
    \resizebox{0.48\textwidth}{!}{
    \begin{tabular}{cccccc}
    \toprule
    \textbf{Tab Formt} & \textbf{TAF} & \textbf{PHT} & \textbf{SCT} & \textbf{TMP} & \textbf{FQA}\\
    \midrule
    Llama2 & 71.8/78.5 & 75.8/66.4 & \underline{64.0/57.0} & 93.6/92.0 & 73.4/37.7 \\
    Llama3 & \underline{76.6/84.7} & \underline{79.2/67.8} & 62.4/55.9 & \underline{94.1/94.4} & \underline{71.8/40.0} \\
    CodeLlama & 67.6/78.0 & \textbf{81.2/66.1} & \textbf{64.6/53.9} & 94.1/91.5 & 76.1/35.7\\
    DeepSeek & \textbf{70.7/79.7} & 72.5/71.3 & 63.5/51.3 & \textbf{95.7/93.9} & \textbf{74.5/38.6} \\
    \midrule
   \textbf{Tool Mkr} & \textbf{TAF} & \textbf{PHT} & \textbf{SCT} & \textbf{TMP} & \textbf{FQA} \\
    \midrule
    Llama2 & 70.2/81.8 & 65.8/60.2 & 53.9/61.5 & 95.7/91.1 & 61.7/31.0 \\
    Llama3 & 75.5/75.4 & 71.1/69.8 & 63.5/52.2 & \textbf{97.9/92.4} & 62.2/38.5 \\
    CodeLlama & \underline{75.5/85.2} & \underline{74.5/71.2} & \textbf{62.9/57.1} & 95.7/91.7 & \underline{68.1/39.8} \\
    Deepseek & \textbf{76.6/84.7} & \textbf{79.2/67.8} & \underline{62.4/55.9} & \underline{94.1/94.4} & \textbf{71.8/40.0} \\
    \bottomrule
    \end{tabular}
    }
    \caption{Results of \model with different backbone modules. The top half uses \texttt{deepseek-code-7b} as the Tool Maker, while the bottom half uses \texttt{Llama3-8b} as the Table Formatter. The best performance is highlighted in bold and the second-best is underlined. Tab Formt is \textit{Table Formatter} and Tool Mkr is \textit{Tool Maker}.}
   \vspace{-0.3cm}
    \label{tab:different modules}
\end{table}

\subsection{Impact of Foundation Models}
\label{sec:ablation study}
% key insights: {highlight the best combination: Llama3+ Deepseek. This is compatible with our intuition. Why? Llama3 is has the best ability for reading long table, and deepseek is pretrained on code has mode ability on creating tools. We believe ... / one possible reason is ...} The combination of Llama3-8b + CodeLLama-7b is the best combination. 

% Min: This section also doesn't have any insight.  This is merely reading from the table.  Can you be more specific?
To explore the optimal module combinations within the \model framework, we explore various pairings of table formatter and toolmaker modules shown in Table~\ref{tab:different modules}. We find that using \texttt{Llama3-8b} as the table formatter and \texttt{DeepSeek-7b} as the tool maker achieves the best average execution rate (76.8) and accuracy (68.6). This aligns with our expectations given that \texttt{Llama3-8b} excels in processing long tables while \texttt{DeepSeek-7b}, with its pre-training on code, demonstrates superior capability in tool creation (c.f. Table~\ref{tab:appendix different modules} in Appendix~\ref{append: different backbone models}).

\section{Discussion}
To further explore the usefulness of \model, we focus on the following research questions: 1) What is the performance of \model on out-of-domain (OOD) datasets? (Section~\ref{sec:ood results}); 2) How does \model create and utilize tools? (Section~\ref{sec:tool use}); and 3) What is the quality of the explanations generated by \model in actual user cases? (Section~\ref{sec:case study}).

\subsection{Out-of-Domain Results}
\label{sec:ood results}
\begin{figure}[!t]
    \centering
    \begin{minipage}[t]{\columnwidth}
        \centering
        \begin{minipage}[t]{0.3\columnwidth}  
            \centering
            \includegraphics[width=1.05\columnwidth]{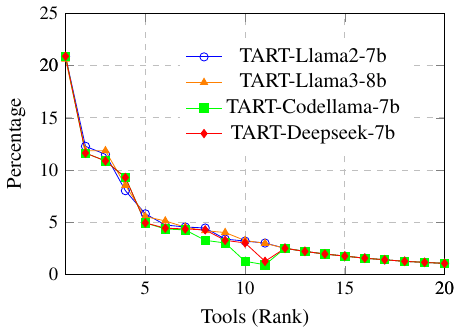}
        \end{minipage}%
        \hfill
        \begin{minipage}[t]{0.35\columnwidth} 
            \centering
            \includegraphics[width=0.75\columnwidth]{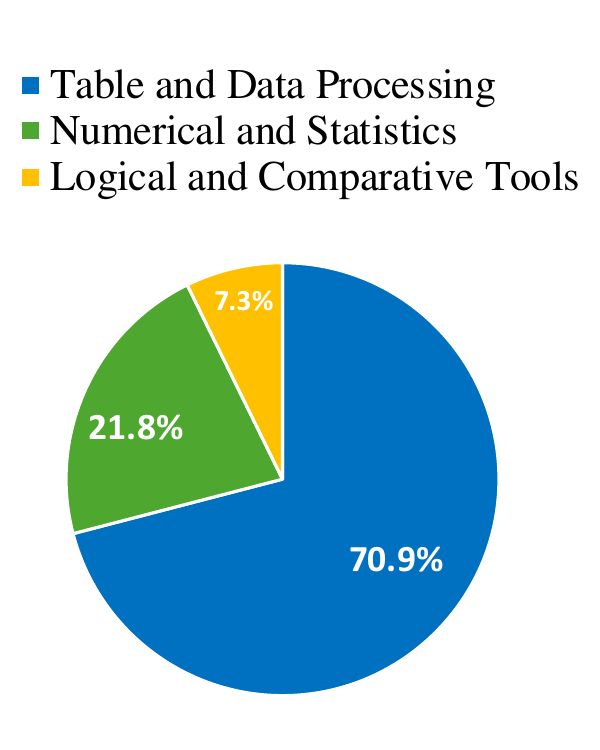}
        \end{minipage}
        % \subcaption{Tool distributions and their categories for the top-20 tools.}
        \label{fig:tool_categories_distributions}
        % \hfill
        \begin{minipage}[t]{0.33\columnwidth}
        \centering
        \vspace{-2.5cm}
        \resizebox{\columnwidth}{!}{  % Resize the table to fit the text width
            % \begin{tabular}{clcl}
            % \toprule
            %      Rank & Tool Name & Rank & Tool Name \\
            %      \midrule
            %      1 & get\_column\_by\_name & 6 & equal\_to \\
            %      2 & get\_column\_cell\_value & 7 & get\_column\_by\_index \\
            %      3 & get\_row\_index\_by\_value & 8 & subtract \\
            %      4 & extract\_price & 9 & divide \\
            %      5 & get\_row\_by\_name & 10 & add \\
            %     \bottomrule
            % \end{tabular}

            \begin{tabular}{cl}
            \toprule
                 Rank & Tool Name\\
                 \midrule
                 1 & get\_column\_by\_name  \\
                 2 & get\_column\_cell\_value \\
                 3 & get\_row\_index\_by\_value \\
                 4 & extract\_price \\
                 5 & get\_row\_by\_name\\
                 6 & equal\_to\\
                 7 & get\_column\_by\_index \\
                 8 & subtract \\
                 9 & divide \\
                10 & add \\
                \bottomrule
            \end{tabular}
        }
        % \subcaption{Top-10 tools in TART \texttt{Codellama-7b}.}
        \label{tab:top10_codellama}
    \end{minipage}
    \end{minipage}
     \vspace{-0.5cm}
    % Table in a separate minipage below
    \caption{Analysis of tool usage in the \model framework. Th left and center panels show the distribution and the categories of the top 20 most-used tools across models. The right shows the top-10 tools in \model \texttt{CodeLlama-7b}.}
    %\vspace{-0.5cm}
    \label{fig:tool_analysis}
\end{figure}

\begin{table}
    \centering
    \fontsize{8}{10}\selectfont
    \setlength{\tabcolsep}{1.3mm}
    \resizebox{0.48\textwidth}{!}{
    \begin{tabular}{clcccc}
    \toprule
         & & \multicolumn{2}{c}{\textbf{TQA}}&  \multicolumn{2}{c}{\textbf{Hybrid TQA}}\\
         \textbf{Model}&\textbf{Setting}& \textbf{ HIT}&  \textbf{WTQ}&  \textbf{TAT}&  \textbf{HYQ} \\
        \toprule
          \multirow{3}{*}{\textbf{Llama2-7b}}& w/ DirectQA & 19.1&23.4&15.4&8.5  \\  
          & w/ CoT & 22.1& 12.7&20.0&6.7  \\
        \rowcolor{lightblue} & w/ TART & 19.2& 17.0& 17.0&6.4 \\
\hdashline
         \multirow{3}{*}{\textbf{Llama3-8b}} & w/ DirectQA & \textbf{51.1}&\textbf{38.8}&20.2&10.1 \\  
        & w/ CoT& 33.8& 26.8& \underline{29.3}& 11.0 \\
        \rowcolor{lightblue}  & w/ TART& \underline{34.6}&32.5& \textbf{29.3} & \textbf{12.2}\\
\hdashline        
         \multirow{3}{*} {\textbf{CodeLlama-7b}}&w/ DirectQA & 17.0&19.1&13.8&6.9 \\  
         & w/ CoT& 16.1&22.6&14.6&9.7 \\
        \rowcolor{lightblue}   & w/ TART&22.3 & 30.3&13.8 & 9.0 \\
\hdashline
         \multirow{3}{*}{\textbf{Deepseek-7b}}& w/ DirectQA & 27.1&26.2&19.7&11.2\\  
         & w/ CoT& 20.5& 26.2&15.3&8.1 \\
        \rowcolor{lightblue}  & w/ TART&29.8& \underline{33.5} & 17.0& \underline{11.2} \\
         \bottomrule
    \end{tabular}
}
    \caption{Out-of-Domain evaluation results for \model framework, highlighting the best (bold) and the second-best (underlined) results.}
   \vspace{-0.3cm}
    \label{tab:ood}
\end{table}

% Table~\ref{tab:ood} 
% Building on the observations from the in-domain datasets presented in Section~\ref{sec: main results}, 
We hypothesize that the \model has enhanced generalization capabilities compared to CoT due to its ability to create and use general tools. To validate this, we further evaluate \model across four different OOD datasets: \texttt{HiTab (HIT)}, \texttt{WikiTableQuestion (WTQ)}, \texttt{TAT-QA (TAT)}, and \texttt{HybridQA (HYQ)}. The results are shown in Table~\ref{tab:ood}. 

The \model method demonstrates variable effectiveness, with notable improvements in certain contexts. For instance, it achieves an average accuracy increase of 29.3\% on the \texttt{WTQ} dataset, indicating robust domain-transfer capabilities. The \texttt{Deepseek-7b} backbone model excels, with 30.6\% increase in accuracy. We hypothesize that this superiority stems from its pre-training on coding tasks, which equips it with the capability of effectively creating and using tools in novel domains, surpassing pure-text-based pretraining models such as \texttt{Llama2-7b}. The analysis in Section~\ref{sec:tool use} supports our hypothesis, suggesting that \model excels in developing generic table reasoning functions that generalize well across various domains.

\subsection{Analysis of Tool Creation}
\label{sec:tool use}

We then performed an in-depth analysis of how \model creates and utilizes tools.

\subsubsection{Tool Distribution.}
\label{subsec:tool distribution}

Figure~\ref{fig:tool_analysis} (top-left) illustrates the tool usage distribution across different backbone models in \model, highlighting a long-tail distribution. The most frequently used tools are primarily associated with table processing (e.g., \texttt{get\_column\_by\_name}) and numerical reasoning (e.g., \texttt{add}), aligning with our observations in Section~\ref{sec: main results}. Figure~\ref{fig:tool_analysis} (middle) provides a detailed breakdown of tool categories for the top 30 tools, showing that table preprocessing and numerical reasoning tools are the most prevalent. This supports the consistency of tool utilization patterns within \model.

%Minipage version:
% \begin{figure}[!t]
%     \centering
%     \begin{minipage}[t]{0.27\textwidth}
%         \centering
%         \includegraphics[width=\textwidth]{figures/tool_overlap_across_models.pdf}
%         \subcaption{Overlapping tool distribution across all datasets in different backbone models between In-Domain datasets and OOD datasets.}
%         \label{fig:tool overlap}
%     \end{minipage}%
%     % \hfill
%     \begin{minipage}[t]{0.24\textwidth}
%         \centering
%         \includegraphics[width=\textwidth]{figures/line_chart_for_rank_1.pdf}
%         \subcaption{The comparison between the number of repeat tools between training set and testing set and correct tools across different backbone models.}
%         \label{fig:rank}
%     \end{minipage}%
%     \caption{Tool Distribution.}
%     \label{fig:tool_distribution}
% \end{figure}

\begin{figure*}[!t]
    \centering
    \includegraphics[width=2.1\columnwidth]{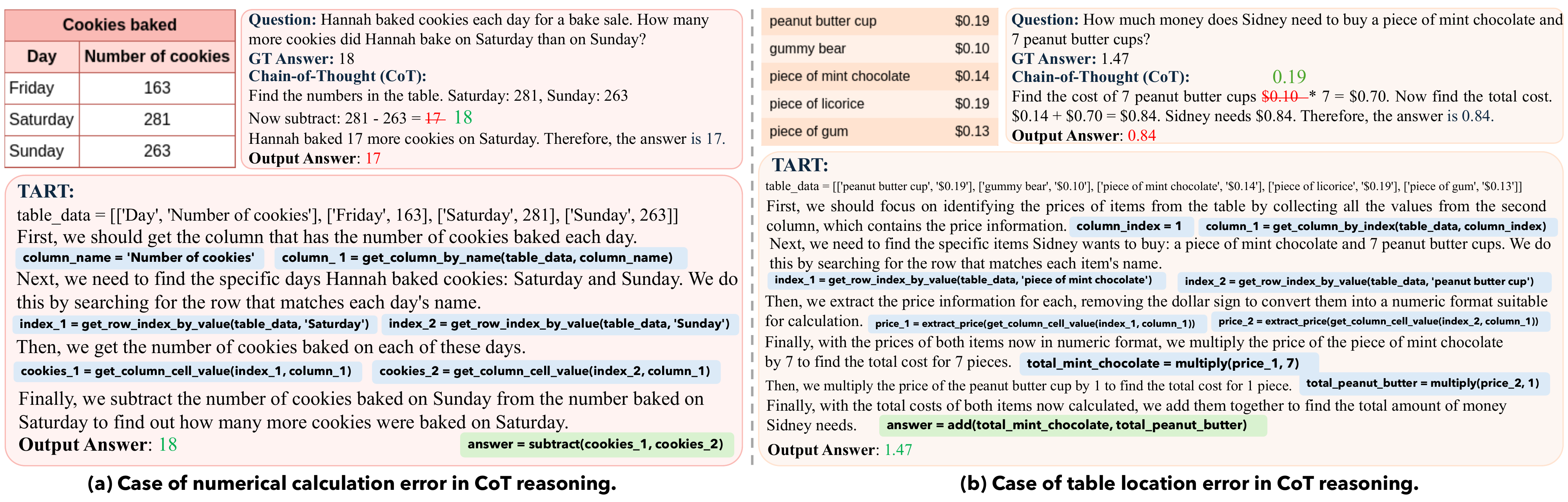}
    \caption{Case Study of \model comparing with CoT reasoning. Panel (a) illustrates a numerical calculation error in CoT where incorrect arithmetic leads to a wrong answer, and panel (b) demonstrates a table location error where CoT fails to retrieve the correct table values. Both errors can be reduced by \model through tool integration.}
    \label{fig:case study}
    \vspace{-0.3cm}
\end{figure*}

\subsubsection{Tool Overlap on OOD Datasets.}
\label{subsec:tool overlap} 
\begin{figure}[!t]
    \centering
    \includegraphics[width=0.9\columnwidth]{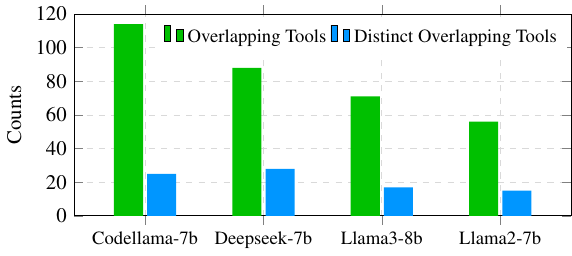}
    \caption{Tool overlap distribution between in-domain and OOD datasets across different backbone models.}
    \label{fig:tool overlap}
   \vspace{-0.3cm}
\end{figure}

% and develop new ones 
Figure~\ref{fig:tool overlap} illustrates the tool overlap between in-domain datasets and OOD datasets. We find that code pre-training models (\texttt{CodeLlama-7b} and \texttt{DeepSeek-7b}) exhibit a tendency to reuse existing tools when adapting to OOD data. However, text pre-training models demonstrate less overlap, indicating that they tend to solve problems by crafting new tools. The tendency to reuse tools might explain why code pre-training models gain better generalization capabilities in unfamiliar data.

\begin{figure}[!t]
    \centering
    \includegraphics[width=0.8\columnwidth]{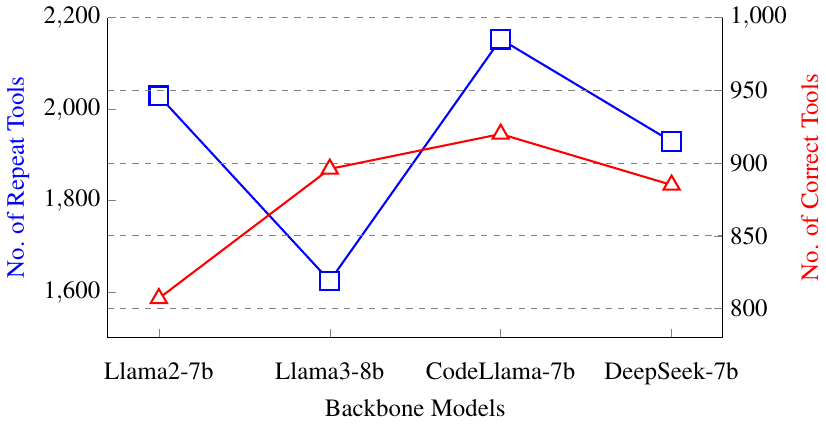}
    \caption{Comparison of the number of repeat tools and correct tools across different backbone models.}
    \label{fig:rank}
   \vspace{-0.3cm}
\end{figure}

\subsubsection{Tool Creation and Usage Analysis.}
% Figure~\ref{fig:rank} further shows that despite \texttt{Llama2-7b} often reuse tools, it also tends to use them in wrong scenarios. In contrast, \textit{CodeLlama-7b} leads in the tendency to reuse tools and the chance of using them correctly. \texttt{Llama3-8b}, while having the least reused tools, performs well in correctly using tools, explaining its better performance. 
Figure~\ref{fig:rank} further reveals that although \texttt{Llama2-7b} frequently reuses tools, it often applies them inappropriately. In contrast, \texttt{CodeLlama-7b} not only exhibits a high rate of tool reuse but also demonstrates a greater accuracy in their appropriate application. Meanwhile, \texttt{Llama3-8b}, despite its lower rate of tool reuse, excels in the correct usage of tools, contributing to its superior performance.

\subsection{Case Study}
\label{sec:case study}
% find two examples of TART better than COT and 1 TART's most common error case, and explain the reason.

To gain deeper insights into the advantages of \model over \texttt{CoT} reasoning, we conducted a case study, shown in Figure~\ref{fig:case study}. The examples highlight the limitation of \texttt{CoT} in numerical reasoning and table preprocessing, such as incorrect calculation in Figure~\ref{fig:case study}(a) and incorrect retrieval inFigure~\ref{fig:case study}(b). Conversely, \model overcomes these challenges effectively via integrating specialized tools like \texttt{subtract} and \texttt{get\_column\_by\_index}. Despite these strengths, \model still encounters issues related to data type mismatches and incorrect programming plans. A detailed analysis of error types in \model can be found in Appendix~\ref{append:error analysis}.

%Conclusion
\section{Conclusion}
We introduce an open-source framework to improve table-based reasoning through the \textit{Tool-Augmented Reasoning framework for Tables (\model)}. This framework solves the challenges of current LLMs' limited ability to understand table structure and execute precise numerical calculations, and maintains explainability. \model consists of a \textit{table formatter} for accurate data representation, a \textit{tool maker} for creating specialized tools, and an \textit{explanation generator} maintaining interpretable explanations. To train \model, we present the \dataset dataset, a novel benchmark containing a diverse set of real-world tables and their tool-augmented solutions. Experiments across nine benchmarks show that integrating our \model method into different open-sourced LLMs enhances accuracy on table-based reasoning.  Furthermore, in-depth analyses reveal that \model effectively learns and uses tools. Future work could extend \model to a multimodal framework by incorporating image-based question-answering and fact-verification to generate richer explanations. Additionally, generating explanations to satisfy the needs of different end users, such as laypersons and experts, could further improve the \model's applicability and impact. 

\section*{Limitations}
\label{limitation}
Despite the promising results, our proposed framework has certain limitations that warrant further investigation:

\paragraph{Computational Complexity.} The \model model may affect efficacy, especially when handling simple questions in quick-response scenarios.

\paragraph{Dataset Coverage.}
While our efforts have focused on expanding the range of our dataset to include a variety of tableQA datasets, some table-related datasets remain unrepresented in \dataset. As a result, despite \model's capacity to adapt to different OOD datasets and tasks, its performance might still different with the complexities and unique challenges of new table tasks and datasets that it has not yet encountered. Having initiated the development of an expansive, versatile tool-enhanced model for tables, we encourage for continued research in this area to further advance the model’s ability to generalize across diverse table configurations.

\section*{Ethics Statement}
\label{ethic_statement}

\paragraph{Transparency and Integrity.}
We ensure that all methodologies, data sources, and technologies used in this study are disclosed transparently. We aim to provide a comprehensive and honest account of our findings, acknowledging both the capabilities and limitations of our proposed solution.

\paragraph{Data Privacy and Security.} 
Our research utilizes datasets that are either publicly available or collected with explicit consent. We adhere to strict data privacy and security protocols to protect the information and ensure it is used solely for the purposes of this research.

\section*{Acknowledgements}
This research is supported by the Ministry of Education, Singapore, under its MOE AcRF TIER 3 Grant (MOE-MOET32022-0001). The computational work for this article was partially performed on resources of the National Supercomputing Centre, Singapore (\url{https://www.nscc.sg}).

\bibliography{custom}
\bibliographystyle{acl_natbib}

\newpage
% \clearpage
\appendix

\section{Dataset Composition for \model Training}
\label{append:training data}
In Table~\ref{tab:dataset comparison}, we show the composition of the seed datasets utilized for training our \model model. These datasets vary in terms of the tasks, the domains, and the types of input and output data. For instance, TabMWP and FinQA focus on TableQA tasks within mathematics and finance domains respectively, requiring a combination of tables, text, and questions as inputs, with short answers as outputs. Meanwhile, PubHealthTab, TabFact, and SCITAB target table fact-checking tasks across health, general Wikipedia, and scientific article domains. These datasets similarly involve tables and claims as inputs but differ in the specifics of the domain-related claims, each producing a short label as an output. 
\begin{table}[!h]
%\centering
\resizebox{0.48\textwidth}{!}{
\begin{tabular}{ccccc}
\toprule
\textbf{Dataset}  & \textbf{Task} & \textbf{Domain}  & \textbf{Input}&\textbf{Output} \\
\midrule
1. TabMWP & TableQA & Maths & Table, Question & Answer (Short)\\
2. FinQA & TableQA  & Finance  & Table, Text, Question & Answer (Short)\\
% 7. TAT-QA~\cite{DBLP:journals/corr/abs-2105-07624}   & TableQA  & Finance  & Table, Text, Question & Answer (Short)\\
3. PubHealthTab & Table Fact Checking & Health & Table, Claim & Label (Short) \\
4. TabFact & Table Fact Checking & Wikipedia  & Table, Claim & Label (Short)  \\
5. SCITAB& Table Fact Checking  & Scientific Articles & Table, Claim & Label (Short) \\ 
\bottomrule                                          
\end{tabular}
}
\caption{Statistics of the seed datasets for \model training, highlighting their respective tasks, domains, and the nature of input and output data.}
\label{tab:dataset comparison}
%\vspace{-0.3cm}
\end{table}

To construct the \dataset, we obtain 11,701, 9,916, and 9,916 training instances for the table formatter $\mathcal{F}$, tool maker $\mathcal{M}$, and explanation generator $\mathcal{E}$, respectively. The detailed statistics is provided in Table~\ref{tab:training samples}.

\begin{table} [!h]
    \centering
    \resizebox{0.48\textwidth}{!}{
    \begin{tabular}{cccccccc}
    \toprule
         % Dataset&  Train&  Dev&  Generated \newline Sample&Executable Sample&Table Formatter&Tool Maker&Explanation Generator\\
             \textbf{Dataset}&  \textbf{Train}&  \textbf{Dev}&  \textbf{Generated} &\textbf{Executable} &\textbf{Table}&\textbf{Tool}&\textbf{Explanation}\\
             &  & & \textbf{Sample}&\textbf{Sample}& \textbf{Formatter}&\textbf{Maker}&\textbf{Generator}\\
         \midrule   TabMWP&23,059&7,686&6,000&5,835&6,000&5,713&5,713\\
         FinQA& 6,251& 883&1,984&1,609&1,967&1,148&1,148\\
         TabFact&92,283&12,792&1,866&1,773&1,866&1,701&1,705\\
         PubHealthTab&1,180&152&1,180&1,075&1,180&958&958 \\
         SciTab&690&-&690&625&688&396&396\\
         Total&123,463&21,513&5,720&10,917&11,701&9,916&9,916\\
    \bottomrule
    \end{tabular}
        }
    \caption{Statistics in dataset \dataset for training \model model. }
    \label{tab:training samples}
\end{table}

\section{Different Backbone Combinations}
\label{append: different backbone models}
In the pursuit of identifying optimal module combinations within the \model framework, we explore various pairings of table formatter and toolmaker modules shown in Table~\ref{tab:appendix different modules}. The combination of \texttt{Llama3-8b} as the table formatter and \texttt{DeepSeek-7b} as the tool maker performs the most effective pairing, having the best average execution rate and accuracy (76.8 and 68.6 respectively). This best combination aligns with our expectations given that \texttt{Llama3-8b} excels in processing long tables while \texttt{DeepSeek-7b}, with its pre-training on code, demonstrates superior capability in tool creation. 
%V2: all results

\begin{table*} [!t]
    \centering
\resizebox{0.9\textwidth}{!}{
    \begin{tabular}{cccccccc}
    \toprule
         \multicolumn{2}{c}{\textbf{Module Name}}&  \multicolumn{3}{c}{\textbf{TableFV}}&  \multicolumn{2}{c}{\textbf{TableQA}}&Avg.\\
         Table Formatter&  Tool Maker&  TabFact&  PubHealthTab&  SCITAB&  TabMWP& FinQA&Exe./Acc.\\
         \midrule
         Llama2&  Llama2& 64.9/79.5& 65.8/59.2& 55.1/60.2& 90.4/91.8&65.4/26.0& 68.3/63.3\\
         Llama2& Llama3& 70.7/75.9& 73.2/65.1& 64.0/46.5&91.0/93.6& 60.6/37.7&71.9/63.8\\
         Llama2&  CodeLlama& 70.2/76.5& \underline{73.8/74.5}& \textbf{64.6/56.5}&94.7/88.8& 71.8/34.1&75.0/66.1\\
         Llama2&  Deepseek& 71.8/78.5& 75.8/66.4& \underline{64.0/57.0}&93.6/92.0& 73.4/37.7&75.7/66.3\\
         \hdashline
         Llama3&  Llama2&  \underline{70.2/81.8}& 65.8/60.2& 53.9/61.5&95.7/91.1& 61.7/31.0&69.5/65.1\\
         Llama3&  Llama3&  75.5/75.4& 71.1/69.8& 63.5/52.2& \textbf{97.9/92.4}&62.2/38.5&74.0/65.7\\
         Llama3&  CodeLlama& 75.5/85.2& 74.5/71.2& 62.9/57.1&95.7/91.7& 68.1/39.8&\underline{75.3/69.0}\\
         Llama3&  Deepseek& \textbf{76.6/84.7}& 79.2/67.8& 62.4/55.9&94.1/94.4& \underline{71.8/40.0}&\textbf{76.8/68.6}\\
         \hdashline
         CodeLlama&  Llama2&64.9/76.2& 69.1/59.2& 53.4/58.9&94.1/89.3& 66.0/26.6&69.5/62.0\\
         CodeLlama&  Llama3& 66.5/71.2& 75.2/69.6& 62.4/57.7&94.1/91.0& 60.1/36.3&71.2/65.2\\
         CodeLlama&  CodeLlama&64.9/75.4& \textbf{77.9/75.0}& 68.5/50.8&95.2/92.2& 71.3/34.3&75.6/65.5\\
         CodeLlama&  DeepSeek&67.6/78.0& 81.2/66.1& 64.6/53.9&94.1/91.5& 76.1/35.7&76.7/65.0\\
         \hdashline
         DeepSeek&  Llama2&  63.3/79.8& 67.1/60.0& 50.0/56.2& 94.7/92.1&63.3/32.8& 67.7/64.2\\
         DeepSeek&  Llama3&  66.5/80.8& 65.1/69.1& 63.5/54.0& 94.1/93.2& 59.6/42.9& 69.8/68.0\\
         DeepSeek&  CodeLlama& 67.0/80.2& 71.1/70.8& 58.4/52.9& 96.8/90.1& 69.7/36.6& 72.6/66.1\\
         DeepSeek&  DeepSeek& 70.7/79.7& 72.5/71.3& 63.5/51.3& \underline{95.7/93.9}&\textbf{74.5/38.6}& 75.4/67.0\\
         \bottomrule
    \end{tabular}
    }
    \caption{The TART framework with different backbone modules,highlighting the best (bold) and the second-best (underlined) results.}
    \label{tab:appendix different modules}
\end{table*}

\section{Tool Use on Different Backbone Models}
\label{append:tool use}
Table~\ref{tab:top10 functions} show the top 10 tools dominate the table processing (e.g., \texttt{get\_column\_by\_name}) and numerical reasoning (e.g., \texttt{add}), consistent with our earlier findings in Section~\ref{sec: main results}. Further illustrating this, Figure~\ref{fig:tool_analysis} (b) presents a tool categorization for the top 30 functions. Table preprocessing tools constitute the highest percentage at 71.0\%, followed by numerical reasoning tools at 21.8\%. Together, these categories account for over 90\% of tool usage, verifying our assumption that \model is better at table preprocessing and numerical reasoning.
\begin{table}[!t]
    \centering
     \resizebox{0.48\textwidth}{!}{
    \begin{tabular}{clll}
    \toprule
         \textbf{Rank} & \textbf{Llama2} & \textbf{Llama3} & \textbf{DeepSeek} \\
         \midrule
         1 & get\_column\_by\_name & get\_column\_by\_name & get\_column\_by\_name \\
         2 & get\_column\_cell\_value & get\_column\_cell\_value & get\_column\_cell\_value \\
         3 & get\_row\_index\_by\_value & get\_row\_index\_by\_value & get\_row\_index\_by\_value \\
         4 & extract\_price & extract\_price & extract\_price \\
         5 & equal\_to & equal\_to & get\_row\_by\_name \\
         6 & get\_column\_by\_index & get\_row\_by\_name & equal\_to \\
         7 & subtract & get\_column\_by\_index & divide \\
         8 & get\_row\_by\_name & divide & get\_column\_by\_index \\
         9 & add & subtract & subtract \\
         10 & multiply & add & add \\
        \bottomrule
    \end{tabular}
    }
    \caption{The top 10 functions across TART \texttt{Llama2-7b}, TART \texttt{Llama3-8b}, and TART \texttt{DeepSeek-7b}.}
    \label{tab:top10 functions}
\end{table}
\section{CoT Baseline Implementation}
%cot baseline 实现：
% 1. 和TART training set的same ID生成一样的cot samples using the prompts with 2 ICL examples, prompts in AppendixXX  by GPT-4
% 2. 和TART 参数一样，训练4个backbone models （llama2, llama3, codellama, deepseek）， with the instructions of "### INSTRUCTION:\nGiven the following table, and question, generate a step-by-step reasoning explanation and the final answer.\n\n\{linearize\_table(sample)\}\n### RESPONSE:\nAnswer: \\{explanation\}\n\n### END"\} 训练时间是个single GPU，12小时，参数和TART一样，i.e., the training process for Llama-2-7b-hf, CodeLlama-7b-hf, and deepseek-coder-7b-instruct-v1.5 requires a single GPU for approximately 12 hours, using a batch size of 4, learning rate of 5e-5, sequence length of 1500, gradient accumulation steps of 2, and 10 training epochs. Training Llama3-8b required up to 2 GPUs for around 12 hours with the same settings.
%Sample Generation:
For a direct and fair comparison with \model, the same number of CoT samples are generated using the same IDs from the \model training dataset. These samples are generated using GPT-4, prompted with two in-context examples (detailed in Appendix~\ref{append:prompt}). In total, we obtain 9,916 training instances. 

%Configuration:
Similar to \model, the CoT baseline was implemented across four different backbone models: \texttt{Llama-2-7b-hf}, \texttt{Llama3-8b}, \texttt{CodeLlama-7b-hf}, and \texttt{DeepSeek-Coder-7b-Instruct-V1.5}. Each model was instructed to generate a step-by-step reasoning explanation followed by the final answer as per the instructions: 
\texttt{INSTRUCTION: Given the following table, and question, generate a step-by-step reasoning explanation and the final answer.}

%Training
The training process was aligned with that of \model to ensure experimental consistency. \texttt{Llama-2-7b-hf}, \texttt{CodeLlama-7b-hf}, and \texttt{deepseek-coder-7b-instruct-v1.5} each requires a single GPU for approximately 12 hours, using a batch size of 4, a learning rate of 5e-5, a sequence length of 1500, gradient accumulation steps of 2, and 10 training epochs. Training \texttt{Llama3-8b} requires up to 2 GPUs for around 10 hours with the same settings.

\begin{figure}[!th]
    \centering
    \includegraphics[width=5.5cm]{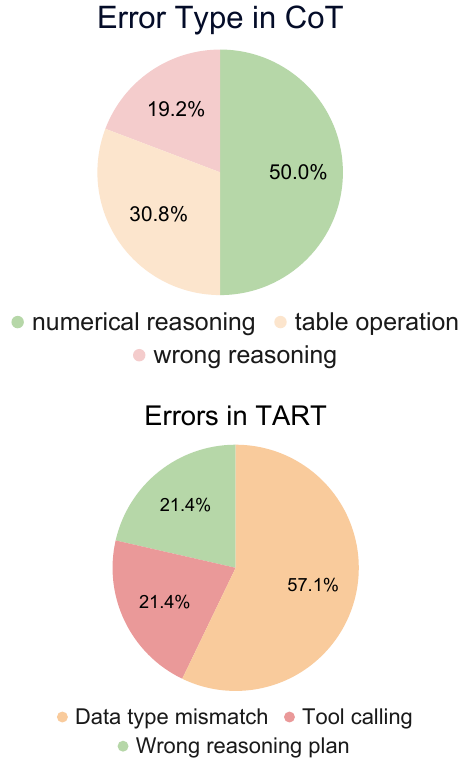}
    \caption{The error types and their distributions of CoT method and our \model framework.}
    \label{fig:error in cot}
     \vspace{-0.7cm}
\end{figure}

\newpage
\section{Error Analysis}
\label{append:error analysis}
% To better observe the error types in CoT, we randomly select 50 CoT error cases and annotate their error types. The result is shown in Figure~\ref{fig:error in cot}. We find that the largest error types in CoT is the wrong numerical reasoning, and then followed by table operation errors. These error analysis verifies our need for proposing \model as our \model solves these challenges by embedding the numerical tools and table operation tools.

To precisely categorize error types in CoT reasoning and \model, we annotate 50 randomly selected error cases for each method. The results (Figure~\ref{fig:error in cot}) shows that the major error type is incorrect numerical reasoning, followed by errors related to table operations. This analysis verifies the necessity for our proposed \model, which addresses these issues by integrating specialized numerical and table operation tools.
% \appendix
\section{Prompts}
\label{append:prompt}
We provide detailed prompts of the \model framework, including the tool discovery process and explanation generation process.
% \paragraph{Tool Discovery Prompts}
\begin{figure}
\begin{quote}
\textbf{Tool Discovery Prompt:}\\
\texttt{Task Description: Given a table and a question, \\the task is to generate a python program to \\answer the question. \\
Requirements:\\
1. First define some functions to be used in the program.\\
2. Try to reuse the functions defined in the previous problems if possible.\\
3. When defining a new function, make sure this function is general enough to be used in other problems.\\
4. Define a function called solution(table\_data) that takes the table data as input and returns the answer to the question. \\
------ \\
'''}\\
\texttt{Table: \colorbox{gray}{\color{white}{Table Content}}}\\
\texttt{Question: \colorbox{gray}{\color{white}{Question}}}\\
\texttt{Answer: \colorbox{gray}{\color{white}{Answer}}}\\
\texttt{'''}\\
\texttt{
table\_data = \colorbox{gray}{\color{white}{table data array}}\\
\\
\#FUNCTION1 Description\\
def FUNCTION1():\\
\colorbox{gray}{\color{white}{Function Body}}\\
\\
\#FUNCTION2 Description\\
def FUNCTION2():\\
\colorbox{gray}{\color{white}{Function Body}}\\
...\\
\\
def solution(table\_data):\\
\colorbox{gray}{\color{white}{Solution Body}}\\
return answer\\
\\
print(solution(table\_data))\\
}
\texttt{------}\\
\texttt{[[FUNCTION\_SOLUTION]]}
\end{quote}
\label{append:tool discovery prompt}
\end{figure}

% \paragraph{Explanation Generation Prompts}
%\label{append:explanation generator prompt}
\begin{figure}
\begin{quote}
\textbf{Explanation Generation Prompt:}\\
\texttt{Task: Transform Python code used for a table question answering task into an easily understandable explanation in natural language embedded with function calls. \\
Follow these requirements: \\
1. The explanation should be the natural language combined with bracketed segments <<< >>> for code.\\
2. The code segments in the brackets <<< >>> should indicate the line number of the code, with the format: \#\#\#<line number>.\\
3. Multiple lines of codes are separated with ';;;' in the brackets <<< >>>.} \\
\texttt{------}\\
\texttt{'''}\\
\texttt{Table: \colorbox{gray}{\color{white}{Table Content}}}\\
\texttt{Question: \colorbox{gray}{\color{white}{Question}}}\\
\texttt{Answer: \colorbox{gray}{\color{white}{Answer}}}\\
\texttt{'''}\\
\texttt{Python Code:}\\
\texttt{table\_data = \colorbox{gray}{\color{white}{table data array}}}\\
\\
\texttt{def solution(table\_data):\\
\colorbox{gray}{\color{white}{Line 1 \#\#\#1}}\\
\colorbox{gray}{\color{white}{Line 2 \#\#\#2}}\\
\colorbox{gray}{\color{white}{...}}\\
\colorbox{gray}{\color{white}{Line 5 \#\#\#5}}\\
return answer\\
\\
print(solution(table\_data))}\\
\\
\texttt{Output Explanation:}\\
\texttt{\colorbox{gray}{\color{white}First, we should get the column}}\\
\texttt{\colorbox{gray}{\color{white}that ... <<<\#\#\#1 ;;; \#\#\#2>>>.}}\\
\colorbox{gray}{\color{white}{...}}\\
\texttt{\colorbox{gray}{\color{white}{Finally, we find that <<<\#\#\#5>>>.}}\\}
\texttt{------}\\
\texttt{[[OUTPUT\_EXPLANATION]]}
\end{quote}
\end{figure}

%CoT prompt
\begin{figure}
\begin{quote}
\textbf{CoT Prompt:}\\
\texttt{Task Description: Given a table and a question, the task is to generate a step-by-step reasoning explanation and the final answer.
} \\
\texttt{------}\\
\texttt{```}\\
\texttt{Table: \colorbox{gray}{\color{white}{Table Content}}}\\
\texttt{Question: \colorbox{gray}{\color{white}{Question}}}\\
\texttt{Answer: \colorbox{gray}{\color{white}{Answer}}}\\
\texttt{```}\\
\texttt{Python Code:}\\
\texttt{table\_data = \colorbox{gray}{\color{white}{table data array}}}\\
\\

\texttt{Output Explanation:}\\
\texttt{\colorbox{gray}{\color{white}To answer this question, first, we ... }}\\
\texttt{\colorbox{gray}{\color{white}Second, to determine..., we compare ...}}\\
\colorbox{gray}{\color{white}{...}}\\
\\
\texttt{\colorbox{gray}{\color{white}{Therefore, the answer is ...}}\\}
\texttt{------}\\
\texttt{[[OUTPUT\_EXPLANATION]]}
\end{quote}
\end{figure}

%GPT-4 Upperbound prompt for Table Formatter
\begin{figure}
\begin{quote}
\textbf{TART (GPT-4) Prompt for Table Formatter:}\\
\texttt{Task Description: Given the following table, context and question, format the table into a python array.} \\
\texttt{------}\\
\texttt{```}\\
\texttt{Table: \colorbox{gray}{\color{white}{Table Content}}}\\
\texttt{Question: \colorbox{gray}{\color{white}{Question}}}\\
\texttt{Answer: \colorbox{gray}{\color{white}{Answer}}}\\
\texttt{```}\\
\texttt{Python Code:}\\
\texttt{table\_data = \colorbox{gray}{\color{white}{table data array}}}\\
\\
\texttt{------}\\
\texttt{[[LINEARIZED\_TABLE]]}
\end{quote}
\end{figure}

%GPT-4 Upperbound prompt for Tool Maker
\begin{figure}
\begin{quote}
\textbf{TART (GPT-4) Prompt for Tool Maker:}\\
\texttt{Task Description: Given the following table, context and question, the table\_data, generate the python code to solve it.}\\
\texttt{------} \\
\texttt{```}\\
\texttt{Table: \colorbox{gray}{\color{white}{Table Content}}}\\
\texttt{Question: \colorbox{gray}{\color{white}{Question}}}\\
\texttt{Answer: \colorbox{gray}{\color{white}{Answer}}}\\
\texttt{
table\_data = \colorbox{gray}{\color{white}{table data array}}\\}
\texttt{```}\\
\\
\texttt{Python Code:}\\
\texttt{
\#FUNCTION1 Description\\
def FUNCTION1():\\
\colorbox{gray}{\color{white}{Function Body}}\\
\\
\#FUNCTION2 Description\\
def FUNCTION2():\\
\colorbox{gray}{\color{white}{Function Body}}\\
...\\
\\
def solution(table\_data):\\
\colorbox{gray}{\color{white}{Solution Body}}\\
return answer\\
\\
print(solution(table\_data))\\
}
\texttt{------}\\
\texttt{[[FUNCTION\_SOLUTION]]}
\end{quote}
\end{figure}

\end{document}